\documentclass{article}

\usepackage{arxiv}

\usepackage[utf8]{inputenc} 
\usepackage[T1]{fontenc}    
\usepackage{hyperref}       
\usepackage{url}            
\usepackage{booktabs}       
\usepackage{amsfonts}       
\usepackage{nicefrac}       
\usepackage{microtype}      
\usepackage{lipsum}		
\usepackage{graphicx}
\usepackage{natbib}
\usepackage{doi}
\usepackage{enumitem}
\usepackage{subcaption}
 \usepackage{amsmath}
\usepackage[table]{xcolor}

\title{Checkup2Action: A Multimodal Clinical Check-up Report Dataset for Patient-Oriented Action Card Generation}

\author{
Sike Xiang \\
Department of Computer Science\\
Durham University\\
\texttt{sike.xiang@durham.ac.uk}
\And
Shuang Chen \\
Department of Computer Science\\
Durham University\\
\texttt{shuang.chen@durham.ac.uk}
\And
Kevin Qinghong Lin \\
University of Oxford\\
\texttt{kevin.qh.lin@gmail.com}
\And
Jialin Yu \\
University of Oxford\\
\texttt{yu.jialin@outlook.com}
\And
Yijia Sun \\
Durham University Business School\\
Durham University\\
\texttt{yijia.sun@durham.ac.uk}
\And
Philip Torr \\
University of Oxford\\
\texttt{philip.torr@eng.ox.ac.uk}
\And
Amir Atapour-Abarghouei$^*$ \\
Department of Computer Science\\
Durham University\\
\texttt{amir.atapour-abarghouei@durham.ac.uk}
}

\renewcommand{\shorttitle}{Checkup2Action}

\begin{document}
\maketitle
\begin{abstract}
	Routine clinical check-up reports combine laboratory measurements, physiological assessments, imaging findings and visually structured information, but rarely tell patients what to do next. Translating them into follow-up actions requires models to connect evidence across pages, tables and modalities, identify clinically relevant issues and communicate next steps without unsupported diagnostic or treatment claims. Yet this report-to-action capability remains poorly benchmarked. We introduce \textbf{C2A}, a dataset and benchmark for generating structured \textit{Action Cards} from multimodal check-up reports, together with \textbf{Checkup2Action}, a constrained workflow for the task. C2A contains 2,000 de-identified real-world reports covering physical examinations, laboratory tests, cardiovascular assessments and imaging evidence. Each card specifies one issue, its priority, recommended department, follow-up window, patient-facing explanation and questions for clinicians. Our evaluation measures issue coverage and precision, priority consistency, department and timing accuracy, action complexity, usefulness, readability and safety. Experiments across general-purpose and medical large language models show that no model performs best on every dimension. Clinical experts judged 84\% of evaluated outputs fully reasonable and effective. Removing safety constraints increased problem recall from 0.527 to 0.825, but produced more diagnostic overstatement. C2A therefore supports systematic study of the balance between coverage, actionable guidance and safe patient communication.
\end{abstract}

\keywords{Dataset and Benchmark \and Medical AI Agent}

\section{Introduction}
\label{sec:intro}

Routine clinical check-ups generate reports that clinicians use to review potential health risks and decide whether further examination, referral or monitoring is warranted~\cite{Araujo2025PeriodicHealthExams, Liss2021GeneralHealthChecks}. The same reports are also returned to patients, often without an equally clear account of what should happen next. These documents combine visual layouts, laboratory tables, numerical biomarkers, abnormality flags, specialised symbols and imaging findings. Relevant evidence may be distributed across multiple pages, structured examination blocks, scanned regions and imaging summaries. Understanding what deserves attention therefore requires more than reading isolated sentences. Patients must connect measurements with reference ranges, interpret visual markers and relate findings across report sections. For people without medical training, this creates an ``interpretability gap'' between multimodal check-up evidence and concrete patient action~\cite{gap1, VanDerMee2024LabResultsFormats, Petrovskaya2023PortalTestResultsScopingReview, gap3}.

\begin{figure*}[t]
  \centering
  \includegraphics[width=\textwidth]{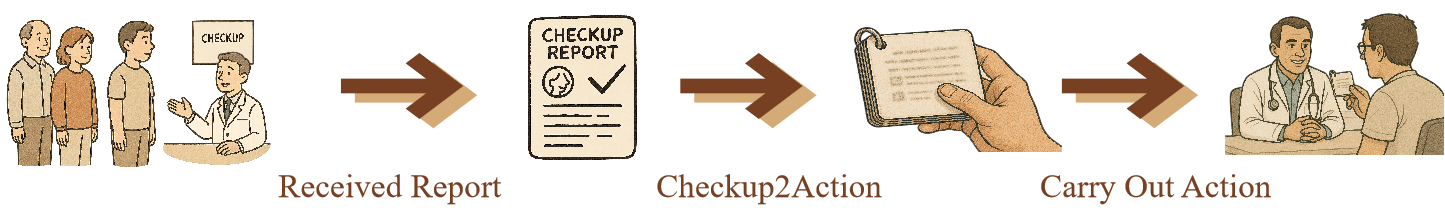}
  \caption{Motivation and task overview. Patients receive multimodal check-up reports containing measurements, abnormality flags and imaging findings, but may remain uncertain about the appropriate next steps. Checkup2Action organises this evidence into prioritised, patient-facing Action Cards for subsequent clinical consultation.}
  \label{fig:1}
\end{figure*}

Patients often begin by scanning visually prominent cues, including ``abnormal'' labels, $\uparrow/\downarrow$ symbols, positive test indicators, highlighted reference-range violations and concluding phrases in laboratory or imaging sections. Visual prominence, however, does not necessarily reflect clinical importance. The same abnormal flag may indicate a negligible deviation in one context and a finding that warrants timely follow-up in another~\cite{abnormal, Timbrell2024ReferenceIntervalLimitations}. Prior work shows that people with limited health literacy can struggle to interpret laboratory values, radiology conclusions and phrases such as ``correlate clinically'' or ``mildly elevated''~\cite{i5, gap3}. This may lead patients to focus excessively on isolated values or overlook findings that deserve further attention. Even when a report includes an ``overall conclusion'' or ``summary opinion'', the content is usually descriptive rather than action oriented. Patients may still be uncertain about which issue to prioritise, which department to consult and when to seek follow-up~\cite{i7}.

Large language models have shown promise in clinical summarisation, patient-facing report simplification and triage assistance~\cite{gap3, masanneck2024_triage_llms}. Most existing systems, however, produce free-form explanations or broad acuity labels rather than structured plans for subsequent action. They seldom assess whether a model can connect evidence across a complete multimodal report and assign each relevant issue an explicit priority, department and follow-up window~\cite{bluethgen2025agenticsystemsradiologydesign}. Existing evaluations also rely heavily on text similarity or small-scale expert ratings. These approaches do not fully capture whether generated outputs identify the right issues, organise them appropriately and remain safe for patient-facing communication~\cite{tam2024frameworkhumanevaluationlarge}. This motivates our central research question: can a model reliably convert a multimodal check-up report into structured and safe actions that help patients prepare for clinical follow-up?

To study this question, we construct \textbf{C2A (Checkup2Action)}, a dataset and benchmark for multimodal check-up-to-action generation (Section~\ref{sec:dataset}). C2A contains 2,000 de-identified real-world check-up reports with physician-reviewed reference Action Cards. It preserves complete reports rather than isolated examination items, requiring models to connect evidence across pages, sections and modalities. We evaluate the generated cards through complementary structured and patient-facing criteria. The structured metrics assess issue coverage and precision, priority consistency, department accuracy, follow-up timing and action complexity. The subjective evaluation measures problem relevance, safety, usefulness, clarity and tone. We further use clinical expert assessment to examine whether the automatic evaluation reflects clinically meaningful output quality.

Alongside the benchmark, we present \textbf{Checkup2Action}, a constrained baseline workflow for generating structured Action Cards (Section~\ref{sec:checkup2action}). The workflow parses the report, organises evidence into clinical sections and produces an ordered set of cards. Each card focuses on one issue and specifies its priority, recommended department, follow-up window, patient-facing explanation and questions for a clinician. The system is constrained to reorganise existing findings into next-step guidance without introducing new diagnostic labels or treatment plans. Fig.~\ref{fig:1} illustrates the intended workflow from a multimodal report to a concise set of actions for subsequent consultation.

Our primary contributions are as follows:

\begin{enumerate}[label=(\roman*)]
\item We introduce \textbf{C2A}, a multimodal clinical check-up report dataset and benchmark containing 2,000 de-identified real-world reports with physician-reviewed annotations for Action Card generation.
\item We formulate check-up-to-action generation as a structured multimodal report understanding task and provide an evaluation protocol covering issue identification, prioritisation, department and timing recommendations, output complexity, patient-facing quality and safety.
\item We provide \textbf{Checkup2Action} as a constrained baseline and conduct controlled comparisons across model families, input representations and safety settings, together with clinical expert validation. The results reveal a consistent tension between broader issue coverage and safe, appropriately calibrated guidance.
\end{enumerate}

\section{Related Work}

We first review work on health check-up reports and medical datasets (Section~\ref{sec:lit:checkup}), then discuss medical AI agents for clinical summarisation, patient-facing communication and triage support (Section~\ref{sec:lit:agents}).

\subsection{Health Check-up Reports and Related Datasets}
\label{sec:lit:checkup}

Routine health check-ups combine several forms of clinical evidence, including vital signs, laboratory tests, functional assessments and imaging examinations. Although examination packages vary across healthcare settings, they commonly include cardiometabolic indicators such as blood pressure, cholesterol, adiposity measures and glucose-related testing~\cite{Araujo2025PeriodicHealthExams, US_Preventive}. The resulting reports organise numerical measurements, reference ranges, abnormality flags, examination blocks, imaging summaries and free-text conclusions across visually distinct sections~\cite{VanDerMee2024LabResultsFormats}. Structured templates and standardised terminology can improve documentation consistency, but consistency for clinicians does not necessarily ensure clarity for patients~\cite{ESR2023StructuredReportingUpdate}. Studies of patient portals show that both patients and healthcare providers often require further explanation to interpret test results and decide on appropriate follow-up~\cite{Petrovskaya2023PortalTestResultsScopingReview}. Reference ranges alone are insufficient because their meaning depends on clinical context and the characteristics of the tested population~\cite{Timbrell2024ReferenceIntervalLimitations}.

Existing datasets address individual parts of this problem. Radiology report simplification supports patient understanding of technical findings~\cite{yang-etal-2023-data}, while MIMIC-CXR pairs chest radiographs with clinical reports for joint image and report modelling~\cite{mimiccxr}. Population resources such as NHANES combine interview data, physical examinations, laboratory measurements and demographic information, but are not organised as report-to-action benchmarks~\cite{zipf2013nhanes}. Lin et al. more recently used real-world health examination reports to adapt and evaluate a preventive health management framework~\cite{lin2026clinically}. These resources support report understanding, generation or population analysis, but they do not assess whether a complete multimodal check-up report can be converted into a prioritised plan for patient follow-up.

Action-oriented clinical NLP provides a closer task setting. Prior work has extracted pending clinical actions from hospital discharge notes~\cite{mullenbach2021clip} and identified recommended tests, their reasons and follow-up intervals in radiology reports~\cite{lau2020followup}. These tasks locate recommendations already expressed in clinical text. C2A instead requires a model to integrate evidence across a complete multimodal report, identify the issues that warrant attention and generate an Action Card for each issue. The cards explicitly represent priority, department and timing while preventing unsupported diagnostic claims or treatment recommendations.

Building such a dataset also requires privacy protection beyond removing names and identification numbers. Medical images may preserve intrinsic biometric characteristics that permit patient re-identification after conventional identifiers have been removed~\cite{packhauser2022reidentification}. Protected health information can also remain in image metadata or in text embedded directly in the pixels~\cite{macdonald2024deidentification}. These risks motivate our decision to release reconstructed, de-identified reports rather than the original clinical PDFs.

\subsection{Medical AI Agents}
\label{sec:lit:agents}

Agentic large language model systems extend text generation by combining instruction following with reasoning, tool use and actions in an external environment. ReAct interleaves reasoning with actions~\cite{ReAct}, while Toolformer studies how language models can learn to invoke external tools~\cite{Toolformer}. Interactive benchmarks and platforms such as OpenHands~\cite{OpenHands}, SWE-agent~\cite{SWE}, Mind2Web~\cite{Mind2Web} and WebArena~\cite{WebArena} further emphasise reproducible evaluation of agents that operate in software or web environments. These studies provide general principles for staged reasoning and tool-mediated workflows, but their tasks do not involve clinical evidence or the safety requirements of patient-facing communication.

In medicine, large language models and agentic systems have been studied for question answering, clinical decision support and documentation generation~\cite{Wang2024}. Clinical summarisation methods condense electronic health records or discharge notes to support clinician review~\cite{Bednarczyk2025}. Patient-facing systems instead rewrite technical medical documents in more accessible language while attempting to preserve the underlying information~\cite{jamanetworkopen}. Related work on triage estimates patient acuity, compares models with emergency medicine professionals or coordinates multiple agents for clinical assessment~\cite{masanneck2024_triage_llms, lu-etal-2024-triageagent}.

These tasks address important parts of clinical communication, but they differ from multimodal check-up action planning. Summarisation explains what a document contains, while triage primarily estimates urgency in acute settings. Checkup2Action must identify relevant findings across a complete check-up report, organise them by priority and attach a suitable department and follow-up window to each issue. It must also present this information to lay users without introducing unsupported diagnoses or treatment plans. Checkup2Action therefore adopts a staged, schema-constrained workflow that connects report parsing, evidence organisation and Action Card generation. The benchmark evaluates the quality and safety of the resulting actions rather than general-purpose agent autonomy.

\section{Dataset and Benchmark}
\label{sec:dataset}

\subsection{C2A Dataset}

We construct \textbf{C2A (Checkup2Action)}, a multimodal dataset for check-up-to-action generation. The research corpus contains 2,000 de-identified, multi-page check-up reports collected through a broader collaboration involving medical and research partners across Asia and Europe. Detailed information about the data sources and contributing institutions will accompany the public dataset. Each report contains evidence from multiple examination items, presented through visual layouts, structured tables, numerical measurements, abnormality flags, embedded images and imaging findings. The reports follow routine documentation practices in the contributing medical institutions and cover a broad range of abnormal findings and health risk indicators.

The study received ethics approval from the hospitals that provided the data and from our university. The hospital review covered non-interventional secondary research using de-identified clinical and health data. The university approval was granted through the relevant departmental or faculty ethics review. Data use is supported by hospital consent materials that permit de-identified clinical data to be used for health data analysis and medical AI research. Approval references and identifying institutional details will be included in the release materials.

All reports were de-identified before dataset construction. Direct identifiers, including names, identification numbers, contact details and exact addresses, were removed or irreversibly transformed. Identifying information was also redacted from report text, page images, scanned content, metadata and visible annotations. Further checks covered filenames, image headers and labels, exact dates, rare combinations of attributes and granular demographic or location information. Reports containing severe or rare findings were reviewed for disclosure risk before release, but this process was not used to select or exclude research samples. The data remained under controlled research governance, and no attempt was made to re-identify any individual.

The original clinical PDFs will not be released. The public dataset will instead contain reconstructed, de-identified reports that preserve the clinical evidence required for Action Card generation. Dates, institution-specific templates, headers, footers, report numbers, barcodes, watermarks and identifiers appearing in image regions will be removed or normalised. The complete de-identification and data use protocol will accompany the public release\footnote{Source code and the reconstructed, de-identified dataset will be publicly available after the review period.}.

Each report covers five broad categories of information. \textbf{Basic Information} includes demographic attributes and vital signs such as age, sex, height, weight and blood pressure. \textbf{General Physical Examination} contains routine assessment findings such as vision measurements and physical signs. \textbf{Laboratory Tests} include standard haematology and biochemical panels, such as blood counts, liver function and renal function tests. \textbf{Imaging Examinations} contain radiological and sonographic findings, including chest X-ray and abdominal ultrasound results. \textbf{Cardiovascular Tests} include cardiac and vascular assessments such as electrocardiography.

C2A is designed as a multimodal document understanding benchmark rather than a text-only summarisation dataset. Each report is represented through its page layout, OCR text, table structure, abnormality markers, units, reference ranges, embedded image regions and imaging findings. Many relevant cues are not expressed in complete sentences. They are conveyed through spatial grouping, table alignment, arrows, highlighted flags and the position of information within a report section. The task therefore requires models to recover and organise evidence from both textual content and visual document structure before generating patient-facing Action Cards.

Reference Action Cards are derived from physician-written report summaries that are not provided to the evaluated models. One physician prepares each report summary, and a second physician checks it. Candidate Action Cards are then drafted from the reviewed summary and examined by a third physician with professional experience in clinical check-ups. The third physician refines the cards before they are used as reference annotations. Model outputs play no role in this process.

The annotation guidelines cover check-up interpretation, structured reporting and patient-facing communication. Physicians determine priority and follow-up windows by considering the full report context, the severity and urgency of each finding, appropriate follow-up intervals and the need for specialist consultation. They also check whether the recommended department and timing are supported by the available evidence and whether the wording avoids unsupported diagnostic or treatment claims. The complete guidelines, adjudication procedure and reliability assessment protocol will be released with the dataset. Each complete report is treated as one sample so that the benchmark preserves document-level context and tests whether models can organise evidence across report sections into coherent Action Cards.

\begin{figure*}[t]
    \centering
    \includegraphics[width=0.95\textwidth]{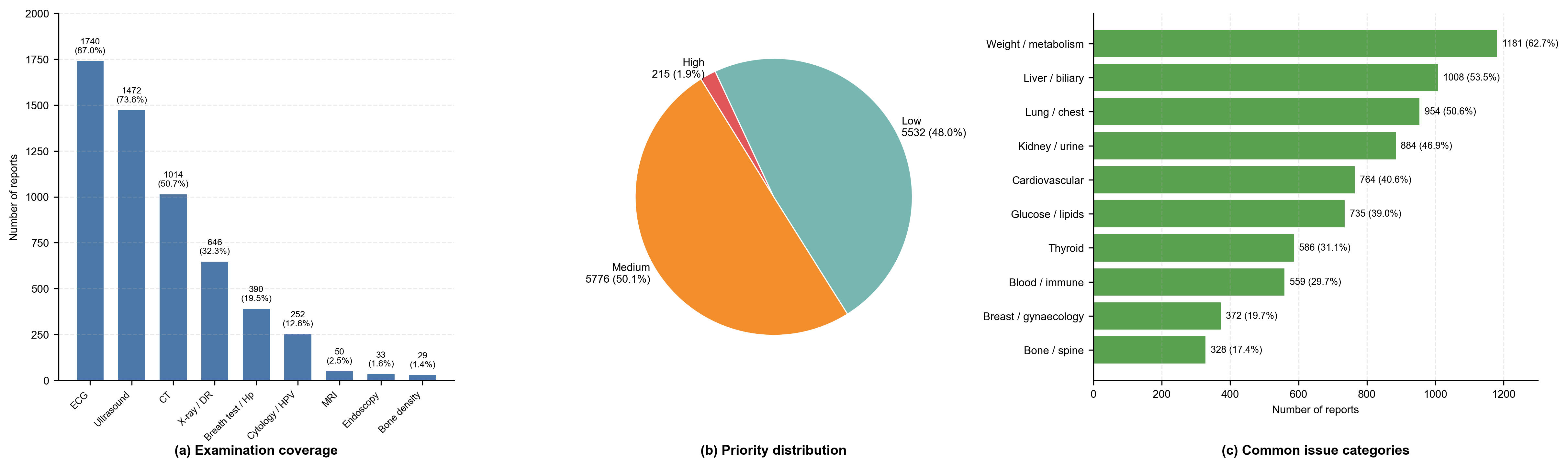}
    \caption{Overview of the C2A dataset.
    (a) Coverage of major examination types.
    (b) Priority distribution of physician-reviewed reference issues.
    (c) Common issue categories derived from physician-written report summaries.}
    \label{fig:dataset}
\end{figure*}

We parse the reports to characterise the composition of C2A. Report length ranges from 2 to 28 pages, with a mean of 12.96 pages. All reports contain core structured sections, including Basic Information and common laboratory panels such as liver function, renal function, glucose, lipids and urinalysis. Additional examination types vary across cases. The dataset contains ECG reports in 1,740 cases, ultrasound in 1,472, CT in 1,014, X-ray or DR in 646, breath testing or Hp testing in 390, cytology or HPV testing in 252, MRI in 50, endoscopy in 33 and bone density examinations in 29 cases, as shown in Fig.~\ref{fig:dataset}(a). This variation reflects routine check-up practice, where additional examinations depend on individual risk factors, clinical indications and screening packages.

Reports with valid reference issues contain between 1 and 24 Action Cards, with a mean of 6.12. The priority distribution is High 1.9\%, Medium 50.1\% and Low 48.0\%, where Low combines the ``long-term focus'' and ``note'' categories. Common issue categories include weight and metabolism, liver and biliary findings, lung and chest findings, kidney and urinary findings, cardiovascular findings, glucose and lipid findings, thyroid findings, blood and immune findings, breast and gynaecological findings, and bone and spine findings. Fig.~\ref{fig:dataset}(b) and Fig.~\ref{fig:dataset}(c) summarise the priority and issue distributions. Detailed statistics and distribution tables will be released with the dataset to support analysis of its scale, difficulty and class imbalance.

\subsection{Benchmark and Evaluation Metrics}
\label{sec:benchmark}

The C2A benchmark evaluates generated Action Cards using two complementary metric families, followed by independent clinical expert validation. Structured metrics assess whether the generated cards identify the reference issues and assign appropriate priorities, departments and follow-up times. Subjective metrics examine whether the complete output is relevant, safe, useful and understandable for lay users. Clinical expert evaluation provides an additional reference for assessing whether the automatic metrics reflect clinically meaningful output quality.

\subsubsection{Structured Consistency}

For report $i$, let $R_i$ and $G_i$ denote the sets of reference and generated issues. We compute the cosine similarity between each pair of issues using BiomedBERT embeddings~\cite{10.1145/3458754}. Let $f_{\text{B}}(\cdot)$ denote the BiomedBERT encoder. The similarity between reference issue $r$ and generated issue $g$ is

\begin{equation}
s(r,g)=\cos\left(f_{\text{B}}(r), f_{\text{B}}(g)\right).
\label{eq:semantic_similarity}
\end{equation}

We perform one-to-one assignment so that each reference or generated issue appears in at most one pair. Assigned pairs with a similarity of at least $\tau=0.8$ are retained in the matched set $M_i$. This threshold follows prior clinical NLP evaluations that treat high cosine similarity as evidence of an acceptable semantic match~\cite{KADHIM2026100895, bioengineering12111194}. \textit{Problem Recall} is defined as

\begin{equation}
\mathrm{Recall}=\frac{\sum_i |M_i|}{\sum_i |R_i|}.
\label{eq:problem_recall}
\end{equation}

Recall measures the proportion of reference issues surfaced by the model. The same matches yield Problem Precision and Problem F1:

\begin{equation}
\mathrm{Precision}
=
\frac{\sum_i |M_i|}{\sum_i |G_i|},
\quad
\mathrm{F1}
=
\frac{2 \cdot \mathrm{Precision}\cdot\mathrm{Recall}}
{\mathrm{Precision}+\mathrm{Recall}}.
\label{eq:problem_precision_f1}
\end{equation}

For matched issues, \textit{Priority Accuracy} measures agreement between the generated priority and the reference priority:

\begin{equation}
\mathrm{Acc}_{\text{pri}}=
\frac{1}{\sum_i |M_i|}
\sum_i \sum_{(r,g)\in M_i} \phi_{\text{pri}}(r,g),
\label{eq:priority_accuracy}
\end{equation}

where $\phi_{\text{pri}}(r,g)=1$ for an exact match, $0.5$ for a predefined near match and $0$ otherwise. \textit{Department Accuracy} and \textit{Time Accuracy} are computed as

\begin{equation}
\mathrm{Acc}_{\text{dep}}
=
\frac{1}{\sum_i |M_i|}
\sum_i \sum_{(r,g)\in M_i}
\mathbf{1}[d_r=d_g],
\quad
\mathrm{Acc}_{\text{time}}
=
\frac{1}{\sum_i |M_i|}
\sum_i \sum_{(r,g)\in M_i}
\mathbf{1}[t_r=t_g].
\label{eq:department_time_accuracy}
\end{equation}

Here, $d$ denotes the recommended department and $t$ denotes the follow-up category, such as immediate, as soon as possible, near term or routine follow-up. \textit{Action Complexity Bias} measures the difference between the number of generated and reference cards:

\begin{equation}
\mathrm{ACB}=\frac{1}{N}\sum_{i=1}^{N}\left(|G_i|-|R_i|\right).
\label{eq:action_complexity_bias}
\end{equation}

A positive value indicates over-generation, while a negative value indicates under-generation. This metric captures whether a system creates too many or too few actions, both of which can reduce the usefulness of patient-facing output.

Recall is not interpreted in isolation because physician-written summaries may use diagnostic labels that Action Cards are explicitly instructed to avoid. We therefore supplement the primary matching procedure with a safety-aware, diagnosis-discounted analysis. Diagnostic labels are mapped to their underlying objective findings when those findings are supported by the source report. Generated cards are also checked for unsupported diagnostic or treatment claims. This analysis distinguishes conservative wording from genuine omission and allows recall to be considered alongside safety and action complexity. Detailed normalisation and penalty rules will accompany the evaluation code.

We also use an LLM judge to assess structured consistency between each matched reference and generated card. The judge considers the issue description, priority, recommended department and follow-up timing in relation to both the reference card and source report. This serves as a complementary validation layer rather than a replacement for embedding-based metrics or physician assessment. Detailed comparisons with the automatic metrics and physician assessments will be released with the evaluation materials.

\subsubsection{Subjective Quality}

We evaluate each report-level output on five subjective dimensions using a scale from 1 to 5. \textit{Problem relevance} measures whether the cards address findings supported by the report rather than offering generic advice. \textit{Safety} assesses whether the output avoids unsupported diagnoses and inappropriate treatment recommendations. \textit{Usefulness} considers whether a lay user can identify and act on the suggested next steps. \textit{Clarity} measures organisation, readability and the visibility of important information. \textit{Tone} assesses whether the language is professional and reassuring without minimising risk or creating unnecessary anxiety.

Outputs are scored through a blinded LLM judging protocol. We use judges from the Gemini, GPT and Claude model families to reduce dependence on any single family and to examine sensitivity to judge selection. Each judge receives the same Action Card JSON and evaluation instructions, without information about the model or workflow that produced the output. Scores are averaged within each dimension. Detailed analyses of agreement among the LLM judges and consistency with physician assessments will be released with the evaluation materials.

\subsubsection{Clinical Expert Evaluation}

We conduct an independent clinical expert evaluation on 500 reports. Three physicians compare the generated Action Cards with the corresponding source reports and assess the appropriateness of the priority, recommended department, follow-up timing and safety of each output. Detailed information about evaluator backgrounds, blinding and disagreement adjudication will be released with the evaluation materials. This evaluation provides an independent clinical reference for assessing the structured metrics and LLM judges.

\section{Checkup2Action}
\label{sec:checkup2action}

\begin{figure*}[t]
    \centering
    \includegraphics[width=\textwidth]{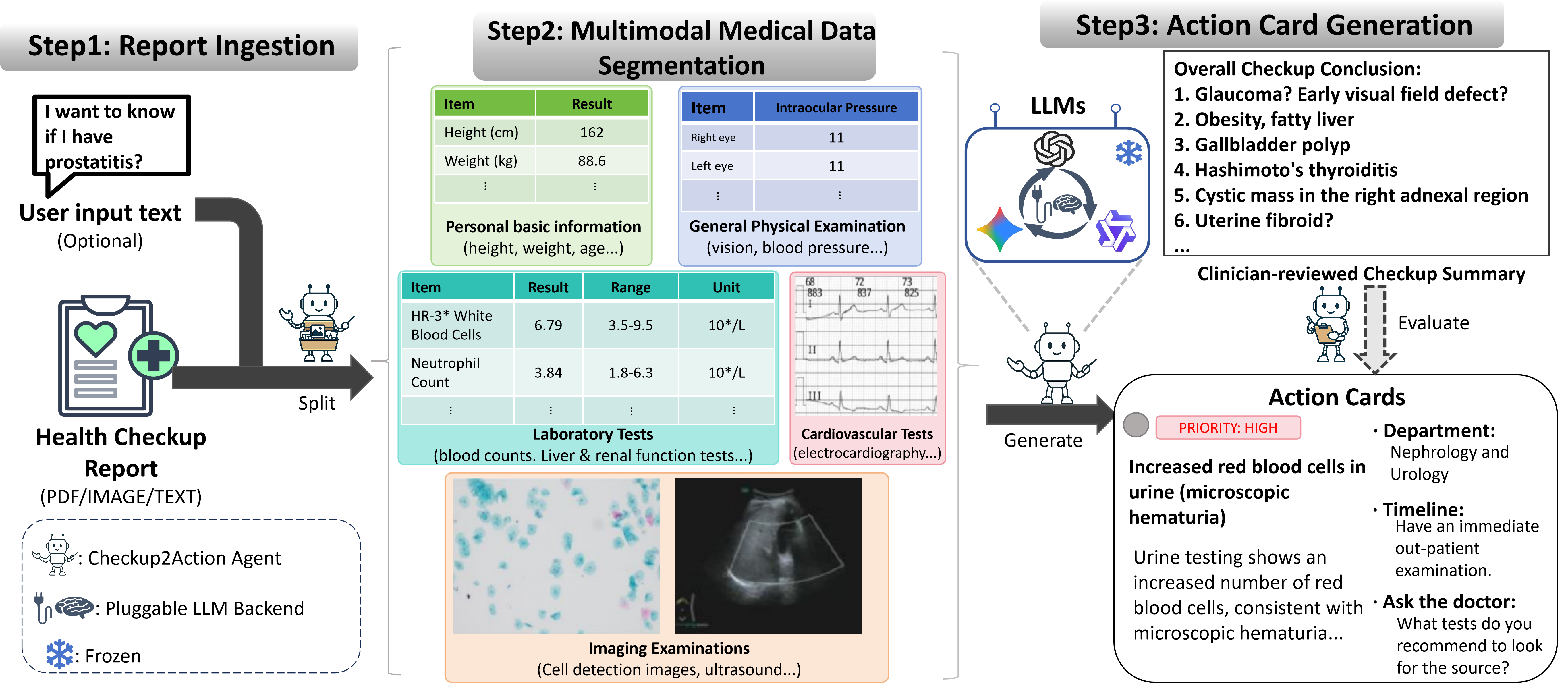}
    \caption{Overview of the Checkup2Action pipeline. A multimodal check-up report is first ingested and parsed, then organised into structured clinical sections covering basic information, physical examinations, laboratory tests, cardiovascular tests, and imaging examinations. The segmented evidence is used to generate prioritised patient-facing Action Cards with recommended department, follow-up timing, and questions for clinicians.}
    \label{fig:overall}
\end{figure*}

The Checkup2Action workflow is shown in Fig.~\ref{fig:overall}. Given a multimodal check-up report, the system outputs an ordered sequence of structured Action Cards. Rather than using a single end-to-end prompt, the workflow separates report ingestion, multimodal section segmentation, schema-constrained generation, and post-generation verification. This decomposition is important for real-world check-up reports, which are often multi-page PDF documents containing visual layouts, free text, numerical tables, reference intervals, abnormality flags, ECG traces, and imaging-related evidence.

At the pipeline level, the system follows three steps. \textbf{Step 1} parses the input report and converts it into a cleaned linearised representation that preserves clinically relevant content while reducing layout noise. \textbf{Step 2} organises the parsed content into structured multimodal sections, such as basic information, physical examination, laboratory tests, cardiovascular tests, and imaging examinations, so that evidence scattered across pages and modalities can be grouped into coherent clinical units. Under the role constraints and output schema described below, \textbf{Step 3} performs schema-constrained generation to produce JSON-formatted Action Cards, followed by schema validation, priority-based ordering, cross-section consistency checking, and safety review.

\subsection{System Pipeline}

\subsubsection{Step 1: Report Ingestion}
Given a check-up report document $d \in \mathcal{D}$, we parse and normalise it into a linearised context:
\begin{equation}
x = \mathcal{L}(d),
\label{eq:linearise}
\end{equation}
where $\mathcal{L}(\cdot)$ denotes a deterministic extraction, cleaning, and normalisation procedure. For PDF reports, we perform page-level parsing and layout extraction to recover heterogeneous report components, including text blocks, table-like regions, abnormality markers, and image objects. The recovered content is then cleaned and linearised into a unified representation while preserving key clinical fields such as item names, measured values, units, reference ranges, abnormal flags, and imaging-related findings. Document artefacts such as headers, footers, duplicated page fragments, and irrelevant decorative elements are removed to reduce downstream noise.

For multimodal models, we provide both the report-page images and the extracted structured representation to preserve the link between visual regions and their recognised textual or tabular content. For text-only models, we use the same extracted representation without page images, which allows controlled comparison between text-only and multimodal settings.

Optionally, a short user-provided summary $u$ can be incorporated by concatenation:
\begin{equation}
x' = [x \,;\, u].
\label{eq:user_concat}
\end{equation}
When no additional summary is provided, we set $x' = x$. The output of this stage serves as the global clinical context for subsequent segmentation and generation.

\subsubsection{Step 2: Multimodal Medical Data Segmentation}
We then organise the linearised content into $m$ structured sections:
\begin{equation}
S = \mathcal{S}(x') = \{s_1, s_2, \dots, s_m\}, \qquad s_j = (t_j, z_j),
\label{eq:sectioning}
\end{equation}
where $t_j$ is the section type, such as basic information, physical examination, laboratory tests, cardiovascular tests, or imaging examinations, and $z_j$ is the corresponding text span or extracted content. Equivalently, $\mathcal{S}(\cdot)$ can be viewed as predicting a set of section boundaries:
\begin{equation}
B = \{(b_j, e_j)\}_{j=1}^{m}, \qquad z_j = x'[b_j:e_j].
\label{eq:boundaries}
\end{equation}

Rather than treating segmentation as a purely free-form generation problem, we adopt a hybrid strategy combining tool-based parsing, rule-based localisation, and LLM verification. First, the parser provides candidate regions from the recovered page structure. Second, rule-based cues such as section headers, table titles, item templates, repeated report patterns, and known examination keywords are used to localise candidate boundaries. Finally, the LLM verifies the available sections and their corresponding page or span ranges, resolving ambiguous cases and refining assignments when evidence from one clinical category is distributed across multiple pages or mixed with other content.

\subsubsection{Step 3: Action Card Generation}
Given the segmented report representation $S$, the model generates a JSON-formatted output string $\hat{y}$ under schema and behavioural constraints:
\begin{equation}
\hat{y}
=
\arg\max_y p_{\theta}
\left(y \mid S,\mathcal{P},\Sigma\right),
\label{eq:constrained_decode}
\end{equation}
where $p_{\theta}$ denotes the large language model and $\Omega$ is the set of valid outputs that satisfy the Action Card schema, field completeness requirements, allowed priority values, and safety constraints that prohibit diagnostic or treatment-prescriptive claims. The model output is then parsed into an ordered sequence of Action Cards:
\[
\mathcal{C} = \{c_1, c_2, \dots, c_n\}.
\]

Each Action Card $c_i$ is a structured object describing one health issue that may require attention, together with a corresponding next-step plan. We model each card as a 6-tuple:
\[
c_i = \bigl(f^{(i)}_1, f^{(i)}_2, \dots, f^{(i)}_6\bigr),
\]
where $f^{(i)}_1$--$f^{(i)}_6$ correspond to \texttt{problem}, \texttt{priority}, \texttt{why}, \texttt{department}, \texttt{time\_suggestion}, and \texttt{questions}, respectively. Here, \texttt{problem} summarises the focal issue, \texttt{priority} takes values in \{``high'', ``medium'', ``long-term focus'', ``note''\}, \texttt{why} provides a short patient-facing explanation of why the issue deserves attention, \texttt{department} specifies a recommended clinical department, \texttt{time\_suggestion} describes the suggested follow-up window, and \texttt{questions} lists questions that the patient may bring to the consultation. This fixed schema keeps the output actionable and comparable across models while limiting redundancy and consultation overload.

\subsection{Behavioural Constraints and Schema}
We constrain the large language model to act as an ``action translator'' for check-up results rather than as a diagnostic or treatment decision maker. Through prompts, schema constraints, and post-generation checks, the model is only allowed to extract, reorganise, and explain information related to next-step actions from existing examination findings and abnormality cues. The generated cards should answer what to do next, when to do it, and which department to consult.

The system is explicitly prohibited from producing definitive diagnostic statements, concrete treatment plans, or drug regimens. For potentially high-risk findings, it must use conservative wording such as ``seek medical care as soon as possible'' or ``follow the current doctor's advice''. These constraints position Checkup2Action as an information reorganisation and action-planning workflow, reducing the risk of over-diagnosis while preserving practical usefulness for patients.

\subsection{Output Parsing and Ordering}
A post-processing module extracts a valid JSON fragment from the raw model output, checks the structural completeness of the \texttt{cards} list and required fields, and sorts the cards according to a predefined priority mapping, for example, ``high'' $>$ ``medium'' $>$ ``long-term focus'' $>$ ``note''. This module does not repair or alter the semantic content of the model output; it only parses, validates, and reorders the generated cards. The resulting ordered \texttt{cards} are then rendered into a user-facing Action Card view for presentation. This design ensures that benchmark comparisons across models are based on a clean and controllable output format.

\subsection{Task Boundary and Clinical Positioning of Action Cards}
Action Cards are designed as a patient-support layer with clearly defined clinical boundaries. Their purpose is not to generate new diagnoses or treatment plans, but to reorganise abnormal findings, risk signals, and follow-up cues already present in check-up reports into structured next-step guidance. In other words, Action Cards are not intended to answer ``What disease does the patient have?'', but rather ``Given the existing check-up results, what should be prioritised next, within what time frame, which department should be consulted, and which questions should be raised?'' Under this framing, Checkup2Action is best understood as a constrained mechanism for multimodal information reorganisation and action translation, rather than as formal clinical decision support or a substitute for diagnosis. Ultimate medical judgement and decision-making remain the responsibility of clinicians.

\section{Experiments}
\label{sec:exp}

\subsection{Experimental Setup}

We evaluate eight models from several commercial and open-weight model families under the same Checkup2Action workflow: \texttt{GPT-5.1}, \texttt{GPT-5-nano}, \texttt{Gemini-3-pro-preview}, \texttt{Gemini-2.5-flash}, \texttt{Claude-sonnet-4.5}, \texttt{Claude-haiku-4.5}, \texttt{Qwen3-VL-8B-Instruct} and \texttt{Grok-4.1-fast}. All models use the same Action Card schema, behavioural constraints and post-processing procedure. Unless otherwise stated, each comparison changes only the model backbone or the input condition. We evaluate the outputs using the structured metrics defined in Section~\ref{sec:benchmark}, the five subjective quality dimensions and clinical expert assessment.

\begin{table*}[t]
\footnotesize
\centering
\setlength{\tabcolsep}{12pt}
\begin{tabular}{l|ccccc}
\toprule
Model & Prob. rec.$\uparrow$ & Prior. acc.$\uparrow$ & Dept. acc.$\uparrow$ & Time acc.$\uparrow$ & Act. compl. bias$|\cdot|\downarrow$ \\
\midrule
\rowcolor{purple!5}
GPT-5.1              & $\underline{0.527}_{0.185}$ & $\underline{0.719}_{0.242}$ & $\textbf{0.844}_{0.237}$ & $\textbf{0.915}_{0.140}$ & $\textbf{-0.447}_{1.794}$ \\
\rowcolor{purple!5}
GPT-5-nano           & $\textit{0.449}_{0.197}$ & $0.712_{0.304}$ & $0.783_{0.278}$ & $0.730_{0.269}$ & $1.544_{2.534}$ \\
\midrule
\rowcolor{blue!5}
Gemini-3-pro-preview & $\textbf{0.602}_{0.191}$ & $0.591_{0.283}$ & $\underline{0.834}_{0.214}$ & $0.807_{0.226}$ & $\underline{1.398}_{2.415}$ \\
\rowcolor{blue!5}
Gemini-2.5-flash     & $0.413_{0.203}$ & $\textit{0.717}_{0.223}$ & $\textit{0.808}_{0.194}$ & $\textit{0.884}_{0.153}$ & $-2.592_{2.447}$ \\
\midrule
\rowcolor{green!5}
Claude-sonnet-4.5    & $0.420_{0.143}$ & $0.698_{0.215}$ & $0.760_{0.185}$ & $\underline{0.890}_{0.166}$ & $-3.000_{2.535}$ \\
\rowcolor{green!5}
Claude-haiku-4.5     & $0.246_{0.124}$ & $\textbf{0.787}_{0.231}$ & $0.791_{0.248}$ & $0.845_{0.240}$ & $-5.922_{3.638}$ \\
\midrule
\rowcolor{orange!5}
Qwen3-VL-8B-Instruct & $0.331_{0.238}$ & $0.650_{0.218}$ & $0.716_{0.288}$ & $0.732_{0.218}$ & $\textit{-1.460}_{3.099}$ \\
\midrule
\rowcolor{gray!8}
Grok-4.1-fast        & $0.408_{0.246}$ & $0.574_{0.364}$ & $0.532_{0.357}$ & $0.662_{0.348}$ & $2.175_{2.911}$ \\
\bottomrule
\end{tabular}
\caption{Structured performance on the Checkup2Action benchmark.
Prob. rec. denotes problem recall, Prior. acc. denotes priority accuracy, Dept. acc. denotes department accuracy, Time acc. denotes time accuracy and Act. compl. bias denotes action complexity bias.
Higher values are better for the first four metrics. For action complexity bias, values with smaller absolute magnitude are better.
Bold, underlined and italicised values indicate the best, second-best and third-best results in each column. Subscripts denote standard deviations.}
\label{tab:checkup2action-structured}
\end{table*}

\subsection{Quantitative Results}

\subsubsection{Structured Metrics}

Tab.~\ref{tab:checkup2action-structured} reports the structured evaluation results. Problem recall must be interpreted in light of the task boundary. Generated Action Cards are instructed to describe findings and subsequent actions without asserting definitive diagnoses. Physician-written reference summaries may use more explicit diagnostic or high-suspicion language. A conservative system may therefore match fewer diagnostically phrased reference issues while producing safer patient-facing output. We consequently consider recall together with priority, department, timing, action complexity and safety. No model performs best across all structured dimensions. \texttt{Gemini-3-pro-preview} achieves the highest problem recall at 0.602, indicating the broadest coverage of reference issues. \texttt{GPT-5.1} reaches a lower recall of 0.527 but obtains the highest department accuracy, time accuracy and absolute action complexity performance. It also ranks second on priority accuracy. These results indicate a stronger balance between issue coverage, action assignment and card count calibration.

The remaining models show different forms of conservative or over-generating behaviour. \texttt{GPT-5-nano} remains close to \texttt{GPT-5.1} on priority accuracy but is weaker on the other structured attributes. \texttt{Claude-haiku-4.5} achieves the highest priority accuracy but also records the lowest recall and the largest negative action complexity bias. It therefore assigns priorities accurately among matched issues but omits many reference issues. \texttt{Gemini-2.5-flash}, \texttt{Qwen3-VL-8B-Instruct} and \texttt{Claude-sonnet-4.5} occupy the middle range on most metrics. \texttt{Grok-4.1-fast} is less accurate on department and timing recommendations. Problem Precision and Problem F1 provide a complementary view of this behaviour. \texttt{GPT-5.1} obtains a precision of 0.662 and an F1 score of 0.614. \texttt{Claude-haiku-4.5} reaches a higher precision of 0.847 but a substantially lower F1 score of 0.371. Its high precision therefore reflects conservative generation rather than broad issue coverage. This comparison shows why precision, recall and action complexity should be considered jointly.

\subsubsection{Subjective Evaluation}

\begin{figure}[t]
    \centering
    \includegraphics[width=0.5\columnwidth]{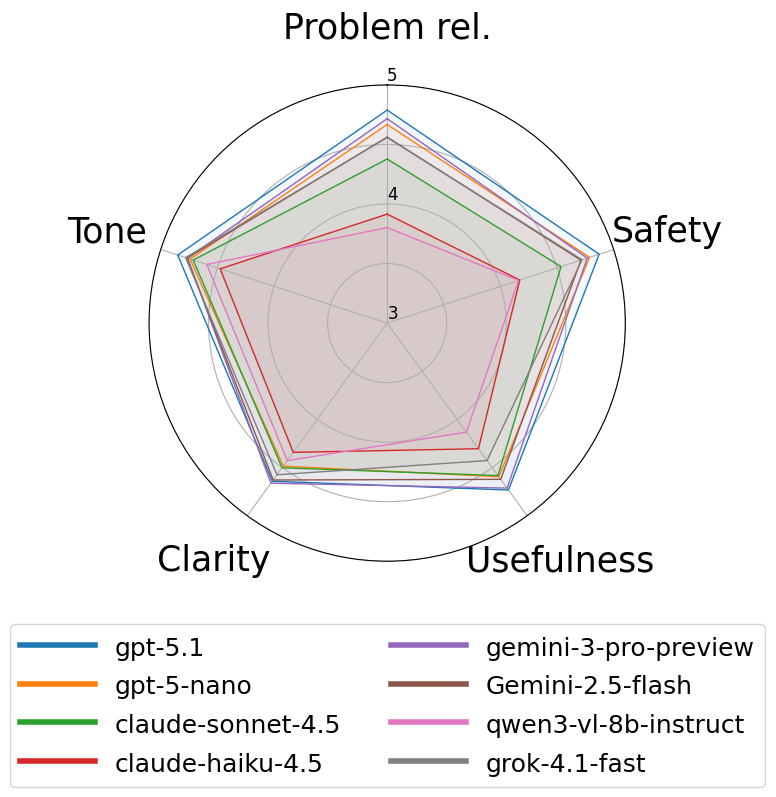}
    \caption{Subjective evaluation across problem relevance, safety, usefulness, clarity and tone. Higher scores indicate better patient-facing quality.}
    \label{fig:checkup2action-radar}
\end{figure}

Fig.~\ref{fig:checkup2action-radar} compares the generated Action Cards across five patient-facing quality dimensions. Scores generally range from 3.8 to 4.9, although the models exhibit different quality profiles. \texttt{GPT-5.1} achieves the strongest overall profile and leads on problem relevance, safety, usefulness and tone. \texttt{Gemini-3-pro-preview} obtains the highest clarity score, while \texttt{Gemini-2.5-flash} also performs well on safety and tone. The Claude models produce relatively safe but information-sparse outputs, consistent with their lower recall and negative action complexity bias in Tab.~\ref{tab:checkup2action-structured}. \texttt{Qwen3-VL-8B-Instruct} receives lower scores on most subjective dimensions. These results reinforce the structured evaluation. Broad issue coverage does not by itself ensure that recommendations are useful, appropriately calibrated or safe for patients.

\subsection{Clinical Expert Validation}

On the 500-report clinical expert evaluation subset, three physicians judged 84\% of the evaluated outputs to be fully reasonable and effective. A further 13\% were considered generally reasonable but differed slightly from the expected follow-up timing. One example was recommending immediate confirmation when confirmation within several days was considered sufficient. The remaining 3\% were judged less appropriate. We also compare the physician assessments with the automatic judges to examine whether automatic evaluation preserves clinically meaningful model comparisons. Detailed agreement analyses will be released with the evaluation materials.

\subsection{Ablation Study}

The ablation experiments examine three questions. First, we test whether visual and layout information improves performance beyond OCR text. Second, we measure the effect of removing the behavioural safety constraints. Third, we compare general-purpose backbones with a medical-domain backbone under the same Checkup2Action workflow.

\begin{figure*}[t]
    \centering
    \begin{subfigure}[t]{0.48\linewidth}
        \includegraphics[width=\linewidth]{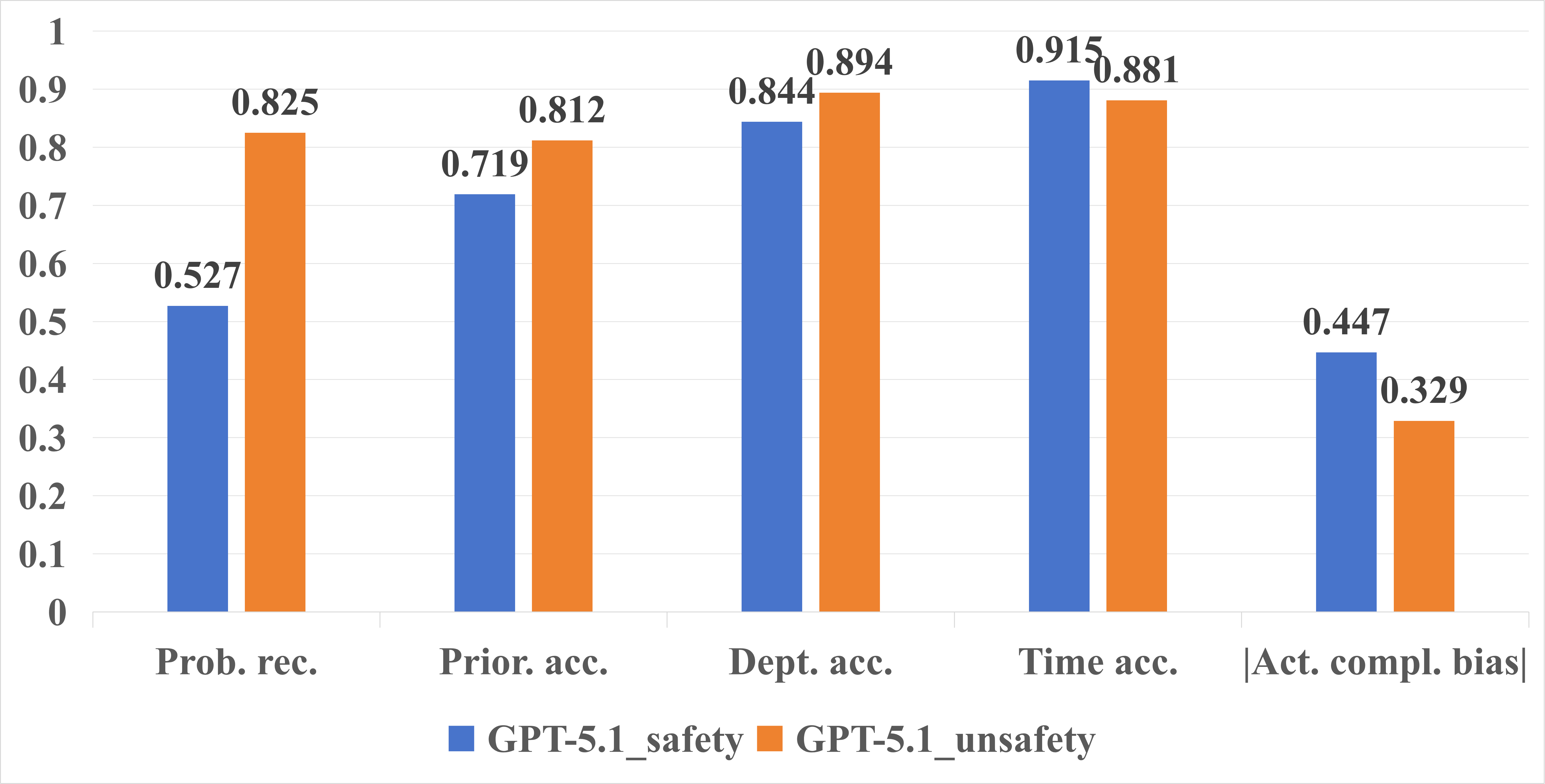}
        \caption{Comparison of safety-constrained and unconstrained variants.}
        \label{fig:compre}
    \end{subfigure}
    \hfill
    \begin{subfigure}[t]{0.48\linewidth}
        \includegraphics[width=\linewidth]{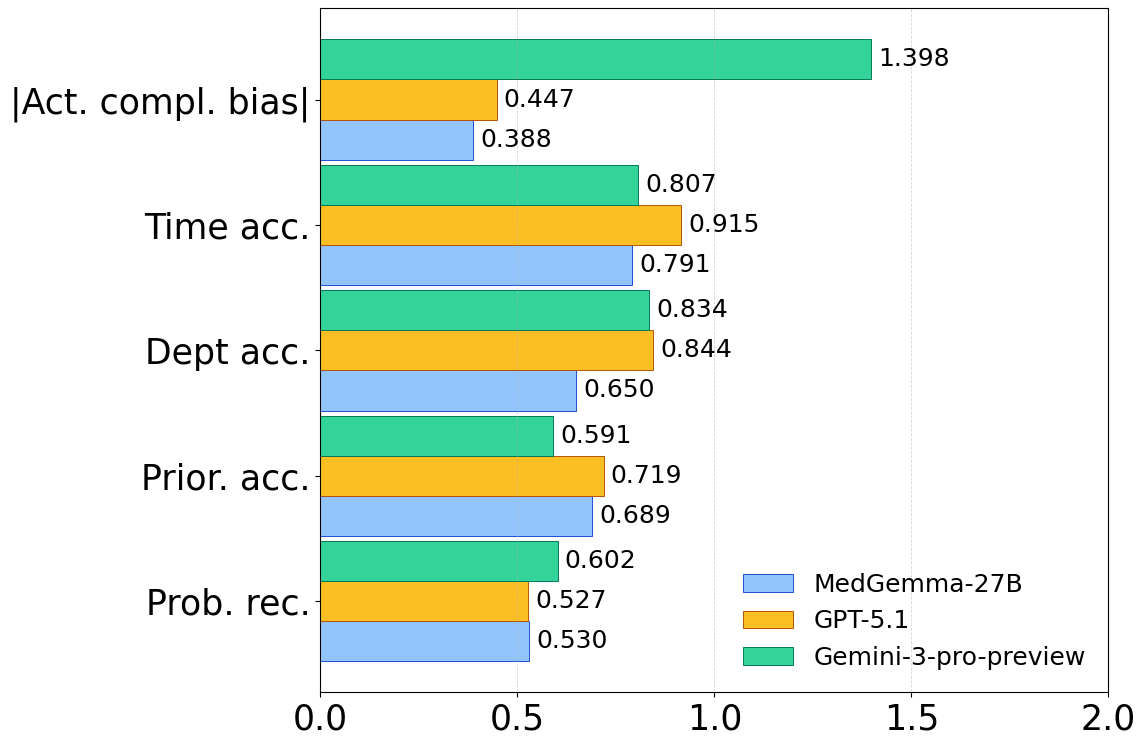}
        \caption{Comparison of \texttt{GPT-5.1}, \texttt{Gemini-3-pro-preview} and \texttt{MedGemma-27B}.}
        \label{fig:compare}
    \end{subfigure}
    \caption{Ablation results for behavioural constraints and model backbone selection.}
    \label{fig:combined_compare}
\end{figure*}

\subsubsection{Input Modality Ablation}

We compare three input settings using \texttt{GPT-5.1}: text-only OCR, layout-aware text and full multimodal input. Layout-aware text preserves table structure, section boundaries, reading order and abnormality flags. Full multimodal input additionally includes the report page images. Relative to full multimodal input, text-only OCR reduces issue coverage by 0.08, priority accuracy by 0.05, department accuracy by 0.06 and timing accuracy by 0.04. Layout-aware text recovers 0.04, 0.02, 0.03 and 0.03 on the same metrics relative to text-only OCR. The recovery from layout-aware text shows that document structure contributes beyond raw OCR. The remaining gap to full multimodal input indicates that page images provide additional information not captured by the extracted text and layout representation.

\subsubsection{Safety Constraint Ablation}
\label{sec:appendix_safety}

We compare the default safety-constrained workflow with a variant in which instructions against definitive diagnoses and treatment recommendations are removed. As shown in Fig.~\ref{fig:compre}, removing the constraints increases problem recall from 0.527 to 0.825. Priority accuracy rises from 0.719 to 0.812 and department accuracy rises from 0.844 to 0.894. Time accuracy decreases from 0.915 to 0.881. These changes indicate that the unconstrained model generates and matches a broader range of issues, but does not improve every downstream action attribute.

\begin{figure*}[t]
  \centering
  \begin{minipage}[b]{0.48\linewidth}
    \centering
    \includegraphics[width=\linewidth]{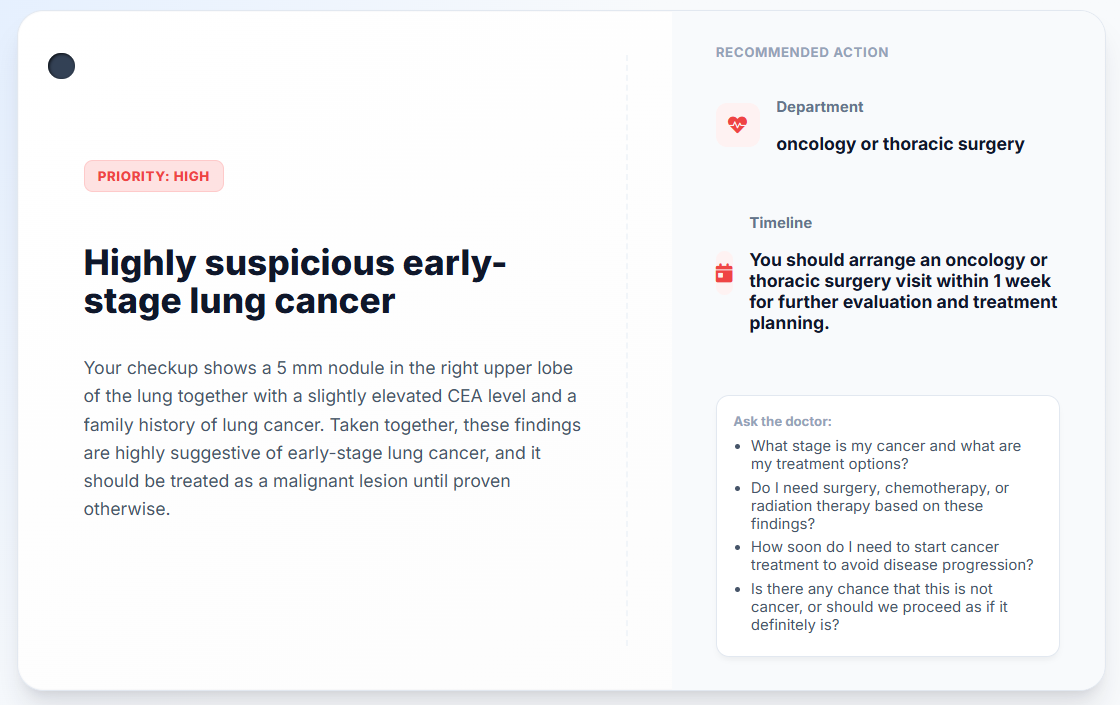}
    {\small \textbf{(a)} Output without safety constraints.}
  \end{minipage}
  \hfill
  \begin{minipage}[b]{0.48\linewidth}
    \centering
    \includegraphics[width=\linewidth]{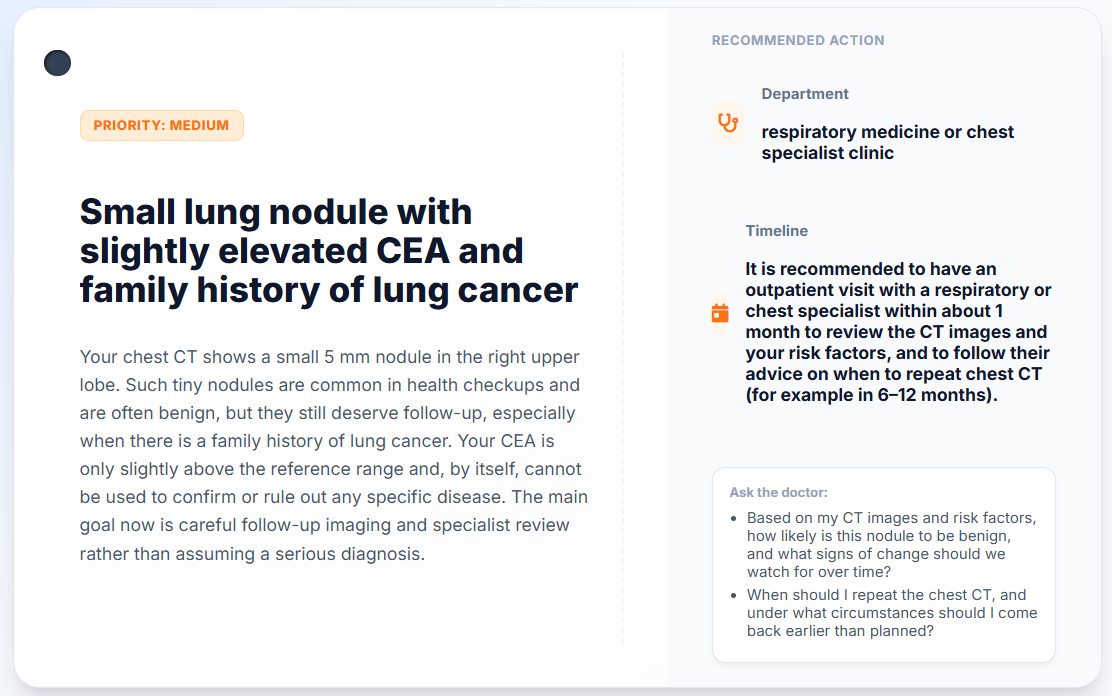}
    {\small \textbf{(b)} Output with safety constraints.}
  \end{minipage}
  \caption{Action Cards generated from the same report without and with behavioural safety constraints.}
  \label{fig:two-figs-vertical}
\end{figure*}

The absolute action complexity bias decreases from 0.447 to 0.329. This means that the number of cards produced by the unconstrained model is slightly closer to the reference count. Action complexity bias measures card count rather than textual concision, so this result does not establish that either variant writes more concise recommendations. The higher recall also comes with weaker safety. The unconstrained variant receives lower Safety and Problem Relevance scores and is more likely to make diagnostic claims such as ``highly suspicious early-stage lung cancer''. Fig.~\ref{fig:two-figs-vertical} compares outputs generated from the same report. The unconstrained output in Fig.~\ref{fig:two-figs-vertical}(a) suggests a specific cancer diagnosis and directs the patient towards staging and treatment. The constrained output in Fig.~\ref{fig:two-figs-vertical}(b) instead reports the observed nodule, mildly elevated CEA and family history. It states that the evidence is insufficient for a diagnosis and recommends follow-up imaging and specialist consultation. The example illustrates the central trade-off. Removing safety constraints improves reference issue coverage but increases diagnostic overstatement.

\subsubsection{Medical Backbone Comparison}

We compare the two strongest general-purpose backbones in the main evaluation, \texttt{GPT-5.1} and \texttt{Gemini-3-pro-preview}, with \texttt{MedGemma-27B}~\cite{medgemma}. The Checkup2Action workflow, behavioural constraints and evaluation protocol remain unchanged. This experiment tests whether replacing a general-purpose model with a medical-domain backbone improves report-to-action generation. As shown in Fig.~\ref{fig:compare}, \texttt{GPT-5.1}, \texttt{MedGemma-27B} and \texttt{Gemini-3-pro-preview} obtain problem recall scores of 0.527, 0.530 and 0.602. \texttt{MedGemma-27B} therefore matches \texttt{GPT-5.1} on issue coverage but performs less well on subsequent action attributes. Its priority, department and time accuracy scores are 0.689, 0.650 and 0.791, compared with 0.719, 0.844 and 0.915 for \texttt{GPT-5.1}. \texttt{Gemini-3-pro-preview} obtains the highest recall but has lower priority accuracy at 0.591 and a larger absolute action complexity bias of 1.398. Medical models are commonly assessed on knowledge and reasoning benchmarks such as MedQA~\cite{MedQA} and PubMedQA~\cite{PubMedQA}, while general-purpose assistants are also optimised for instruction following and conversational alignment~\cite{RLHF}. Checkup2Action requires both medical evidence interpretation and constrained communication for lay users. The results suggest that medical specialisation alone does not guarantee stronger performance on this combination of requirements. They do not isolate model training as the cause because the compared systems also differ in architecture, scale and alignment.

\section{Conclusion}

We introduced \textbf{C2A (Checkup2Action)}, a dataset and benchmark for converting multimodal clinical check-up reports into structured, patient-facing Action Cards, together with a constrained baseline workflow for the task. C2A contains 2,000 de-identified real-world reports and requires models to connect evidence across pages, tables, abnormality markers and imaging findings. Unlike report summarisation, the task assigns each relevant issue a priority, recommended department and follow-up window while preventing unsupported diagnostic or treatment claims. Experiments show that no model performs best across every requirement. General-purpose models provide a stronger overall balance than the evaluated medical-domain backbone, while the input ablation confirms that layout and page images contribute information beyond OCR text. Removing behavioural constraints increases problem recall from 0.527 to 0.825 but also produces more diagnostic overstatement. On the clinical expert subset, 84\% of the evaluated outputs were judged fully reasonable and effective. These findings show that report-to-action generation must be assessed through action correctness, calibration and safety as well as issue coverage.

The current evaluation is retrospective and reflects the report formats and clinical practices represented in C2A. Further work should examine additional hospitals, healthcare systems, languages, measurement conventions and document templates. Prospective studies with patients and clinicians are also needed to determine whether Action Cards improve comprehension and support appropriate follow-up in practice. Dedicated models may further improve the balance between coverage and safety as more annotated multimodal reports become available. C2A provides a foundation for studying how clinical report understanding can support concrete patient actions without crossing the boundary into diagnosis or treatment.

\bibliographystyle{unsrtnat}
\bibliography{references}  

@article{gap1,
    author = {Lazaro, Gerardo},
    title = {When Positive is Negative: Health Literacy Barriers to Patient Access to Clinical Laboratory Test Results},
    journal = {The Journal of Applied Laboratory Medicine},
    volume = {8},
    number = {6},
    pages = {1133-1147},
    year = {2023},
    month = {09},
    issn = {2576-9456},
    doi = {10.1093/jalm/jfad045},
    url = {https://doi.org/10.1093/jalm/jfad045},
    eprint = {https://academic.oup.com/jalm/article-pdf/8/6/1133/52760070/jfad045.pdf},
}

@article{gap3,
  author  = {Sterling, N. W. and Brann, F. and Frisch, S. O. and Schrager, J. D.},
  title   = {Patient-Readable Radiology Report Summaries Generated via Large Language Model: Safety and Quality},
  journal = {Journal of Patient Experience},
  year    = {2024},
  volume  = {11},
  doi     = {10.1177/23743735241259477}
}

@article{abnormal,
    author = {Steitz, Bryan D. and Turer, Robert W. and Lin, Chen-Tan and MacDonald, Scott and Salmi, Liz and Wright, Adam and Lehmann, Christoph U. and Langford, Karen and McDonald, Samuel A. and Reese, Thomas J. and Sternberg, Paul and Chen, Qingxia and Rosenbloom, S. Trent and DesRoches, Catherine M.},
    title = {Perspectives of Patients About Immediate Access to Test Results Through an Online Patient Portal},
    journal = {JAMA Network Open},
    volume = {6},
    number = {3},
    pages = {e233572-e233572},
    year = {2023},
    month = {03},
    issn = {2574-3805},
    doi = {10.1001/jamanetworkopen.2023.3572},
    url = {https://doi.org/10.1001/jamanetworkopen.2023.3572},
    eprint = {https://jamanetwork.com/journals/jamanetworkopen/articlepdf/2802672/steitz_2023_oi_230142_1678732342.1401.pdf},
}

@article{i5,
  author  = {Lustria, Mia Liza A. and Aliche, Obianuju and Killian, Michael O. and He, Zhe},
  title   = {Enhancing patient engagement and understanding: Is providing direct access to laboratory results through patient portals adequate?},
  journal = {JAMIA Open},
  year    = {2025},
  volume  = {8},
  number  = {2},
  pages   = {ooaf009},
  doi     = {10.1093/jamiaopen/ooaf009},
  pmid    = {40130170},
  pmcid   = {PMC11932648}
}

@article{i7,
    author = {Steitz, Bryan D. and Turer, Robert W. and Salmi, Liz and Suresh, Uday and MacDonald, Scott and DesRoches, Catherine M. and Wright, Adam and Louissaint, Jeremy and Rosenbloom, S. Trent},
    title = {Repeated Access to Patient Portal While Awaiting Test Results and Patient-Initiated Messaging},
    journal = {JAMA Network Open},
    volume = {8},
    number = {4},
    pages = {e254019-e254019},
    year = {2025},
    month = {04},
    issn = {2574-3805},
    doi = {10.1001/jamanetworkopen.2025.4019},
    url = {https://doi.org/10.1001/jamanetworkopen.2025.4019},
    eprint = {https://jamanetwork.com/journals/jamanetworkopen/articlepdf/2832287/steitz_2025_oi_250180_1743446461.57741.pdf},
}

@article{masanneck2024_triage_llms,
  author  = {Masanneck, Lars and Schmidt, Linea and Seifert, Antonia and K{\"o}lsche, Tristan and Huntemann, Niklas and Jansen, Robin and Mehsin, Mohammed and Bernhard, Michael and Meuth, Sven G. and B{\"o}hm, Lennert and Pawlitzki, Marc},
  title   = {Triage Performance Across Large Language Models, ChatGPT, and Untrained Doctors in Emergency Medicine: Comparative Study},
  journal = {Journal of Medical Internet Research},
  year    = {2024},
  volume  = {26},
  pages   = {e53297},
  doi     = {10.2196/53297},
  url     = {https://www.jmir.org/2024/1/e53297}
}

@misc{bluethgen2025agenticsystemsradiologydesign,
      title={Agentic Systems in Radiology: Design, Applications, Evaluation, and Challenges}, 
      author={Christian Bluethgen and Dave Van Veen and Daniel Truhn and Jakob Nikolas Kather and Michael Moor and Malgorzata Polacin and Akshay Chaudhari and Thomas Frauenfelder and Curtis P. Langlotz and Michael Krauthammer and Farhad Nooralahzadeh},
      year={2025},
      eprint={2510.09404},
      archivePrefix={arXiv},
      primaryClass={cs.AI},
      url={https://arxiv.org/abs/2510.09404}, 
}

@misc{tam2024frameworkhumanevaluationlarge,
      title={A Framework for Human Evaluation of Large Language Models in Healthcare Derived from Literature Review}, 
      author={Thomas Yu Chow Tam and Sonish Sivarajkumar and Sumit Kapoor and Alisa V Stolyar and Katelyn Polanska and Karleigh R McCarthy and Hunter Osterhoudt and Xizhi Wu and Shyam Visweswaran and Sunyang Fu and Piyush Mathur and Giovanni E. Cacciamani and Cong Sun and Yifan Peng and Yanshan Wang},
      year={2024},
      eprint={2405.02559},
      archivePrefix={arXiv},
      primaryClass={cs.CL},
      url={https://arxiv.org/abs/2405.02559}, 
}

@misc{ReAct,
      title={ReAct: Synergizing Reasoning and Acting in Language Models}, 
      author={Shunyu Yao and Jeffrey Zhao and Dian Yu and Nan Du and Izhak Shafran and Karthik Narasimhan and Yuan Cao},
      year={2023},
      eprint={2210.03629},
      archivePrefix={arXiv},
      primaryClass={cs.CL},
      url={https://arxiv.org/abs/2210.03629}, 
}

@misc{Toolformer,
      title={Toolformer: Language Models Can Teach Themselves to Use Tools}, 
      author={Timo Schick and Jane Dwivedi-Yu and Roberto Dessì and Roberta Raileanu and Maria Lomeli and Luke Zettlemoyer and Nicola Cancedda and Thomas Scialom},
      year={2023},
      eprint={2302.04761},
      archivePrefix={arXiv},
      primaryClass={cs.CL},
      url={https://arxiv.org/abs/2302.04761}, 
}

@misc{OpenHands,
      title={OpenHands: An Open Platform for AI Software Developers as Generalist Agents}, 
      author={Xingyao Wang and Boxuan Li and Yufan Song and Frank F. Xu and Xiangru Tang and Mingchen Zhuge and Jiayi Pan and Yueqi Song and Bowen Li and Jaskirat Singh and Hoang H. Tran and Fuqiang Li and Ren Ma and Mingzhang Zheng and Bill Qian and Yanjun Shao and Niklas Muennighoff and Yizhe Zhang and Binyuan Hui and Junyang Lin and Robert Brennan and Hao Peng and Heng Ji and Graham Neubig},
      year={2025},
      eprint={2407.16741},
      archivePrefix={arXiv},
      primaryClass={cs.SE},
      url={https://arxiv.org/abs/2407.16741}, 
}

@misc{SWE,
      title={SWE-agent: Agent-Computer Interfaces Enable Automated Software Engineering}, 
      author={John Yang and Carlos E. Jimenez and Alexander Wettig and Kilian Lieret and Shunyu Yao and Karthik Narasimhan and Ofir Press},
      year={2024},
      eprint={2405.15793},
      archivePrefix={arXiv},
      primaryClass={cs.SE},
      url={https://arxiv.org/abs/2405.15793}, 
}

@misc{WebArena,
      title={WebArena: A Realistic Web Environment for Building Autonomous Agents}, 
      author={Shuyan Zhou and Frank F. Xu and Hao Zhu and Xuhui Zhou and Robert Lo and Abishek Sridhar and Xianyi Cheng and Tianyue Ou and Yonatan Bisk and Daniel Fried and Uri Alon and Graham Neubig},
      year={2024},
      eprint={2307.13854},
      archivePrefix={arXiv},
      primaryClass={cs.AI},
      url={https://arxiv.org/abs/2307.13854}, 
}

@misc{Mind2Web,
      title={Mind2Web: Towards a Generalist Agent for the Web}, 
      author={Xiang Deng and Yu Gu and Boyuan Zheng and Shijie Chen and Samuel Stevens and Boshi Wang and Huan Sun and Yu Su},
      year={2023},
      eprint={2306.06070},
      archivePrefix={arXiv},
      primaryClass={cs.CL},
      url={https://arxiv.org/abs/2306.06070}, 
}

@article{Wang2024,
  author  = {Wang, Dandan and Zhang, Shiqing},
  title   = {Large language models in medical and healthcare fields: applications, advances, and challenges},
  journal = {Artificial Intelligence Review},
  year    = {2024},
  volume  = {57},
  number  = {11},
  pages   = {299},
  doi     = {10.1007/s10462-024-10921-0},
  url     = {https://doi.org/10.1007/s10462-024-10921-0},
  issn    = {1573-7462}
}

@article{Bednarczyk2025,
  author  = {Bednarczyk, Lydie and Reichenpfader, Daniel and Gaudet-Blavignac, Christophe and Ette, Amon Kenna and Zaghir, Jamil and Zheng, Yuanyuan and Bensahla, Adel and Bjelogrlic, Mina and Lovis, Christian},
  title   = {Scientific Evidence for Clinical Text Summarization Using Large Language Models: Scoping Review},
  journal = {Journal of Medical Internet Research},
  year    = {2025},
  volume  = {27},
  pages   = {e68998},
  doi     = {10.2196/68998},
  url     = {https://doi.org/10.2196/68998},
  issn    = {1438-8871}
}

@article{jamanetworkopen,
    author = {Zaretsky, Jonah and Kim, Jeong Min and Baskharoun, Samuel and Zhao, Yunan and Austrian, Jonathan and Aphinyanaphongs, Yindalon and Gupta, Ravi and Blecker, Saul B. and Feldman, Jonah},
    title = {Generative Artificial Intelligence to Transform Inpatient Discharge Summaries to Patient-Friendly Language and Format},
    journal = {JAMA Network Open},
    volume = {7},
    number = {3},
    pages = {e240357-e240357},
    year = {2024},
    month = {03},
    issn = {2574-3805},
    doi = {10.1001/jamanetworkopen.2024.0357},
    url = {https://doi.org/10.1001/jamanetworkopen.2024.0357},
    eprint = {https://jamanetwork.com/journals/jamanetworkopen/articlepdf/2815868/zaretsky_2024_oi_240032_1709070763.44144.pdf},
}

@inproceedings{lu-etal-2024-triageagent,
    title = "{T}riage{A}gent: Towards Better Multi-Agents Collaborations for Large Language Model-Based Clinical Triage",
    author = "Lu, Meng  and
      Ho, Brandon  and
      Ren, Dennis  and
      Wang, Xuan",
    editor = "Al-Onaizan, Yaser  and
      Bansal, Mohit  and
      Chen, Yun-Nung",
    booktitle = "Findings of the Association for Computational Linguistics: EMNLP 2024",
    month = nov,
    year = "2024",
    address = "Miami, Florida, USA",
    publisher = "Association for Computational Linguistics",
    url = "https://aclanthology.org/2024.findings-emnlp.329/",
    doi = "10.18653/v1/2024.findings-emnlp.329",
    pages = "5747--5764",
}

@misc{MedQA,
      title={What Disease does this Patient Have? A Large-scale Open Domain Question Answering Dataset from Medical Exams}, 
      author={Di Jin and Eileen Pan and Nassim Oufattole and Wei-Hung Weng and Hanyi Fang and Peter Szolovits},
      year={2020},
      eprint={2009.13081},
      archivePrefix={arXiv},
      primaryClass={cs.CL},
      url={https://arxiv.org/abs/2009.13081}, 
}

@inproceedings{PubMedQA,
    title = "{P}ub{M}ed{QA}: A Dataset for Biomedical Research Question Answering",
    author = "Jin, Qiao  and
      Dhingra, Bhuwan  and
      Liu, Zhengping  and
      Cohen, William  and
      Lu, Xinghua",
    editor = "Inui, Kentaro  and
      Jiang, Jing  and
      Ng, Vincent  and
      Wan, Xiaojun",
    booktitle = "Proceedings of the 2019 Conference on Empirical Methods in Natural Language Processing and the 9th International Joint Conference on Natural Language Processing (EMNLP-IJCNLP)",
    month = nov,
    year = "2019",
    address = "Hong Kong, China",
    publisher = "Association for Computational Linguistics",
    url = "https://aclanthology.org/D19-1259/",
    doi = "10.18653/v1/D19-1259",
    pages = "2567--2577"
}

@misc{RLHF,
      title={Learning to summarize from human feedback}, 
      author={Nisan Stiennon and Long Ouyang and Jeff Wu and Daniel M. Ziegler and Ryan Lowe and Chelsea Voss and Alec Radford and Dario Amodei and Paul Christiano},
      year={2022},
      eprint={2009.01325},
      archivePrefix={arXiv},
      primaryClass={cs.CL},
      url={https://arxiv.org/abs/2009.01325}, 
}

@article{Araujo2025PeriodicHealthExams,
  author  = {Araujo, G. C. and Ribeiro, C. B. and Costa, M. C. M. and Evangelista, M. L. P. and Lima, M. F. and De Paula, M. C. and Ferreira, V. L. and Araujo, F. A. G. D. R.},
  title   = {Evidence-Based Periodic Health Examinations for Adults: A Practical Guide},
  journal = {Cureus},
  year    = {2025},
  volume  = {17},
  number  = {3},
  pages   = {e79963},
  doi     = {10.7759/cureus.79963},
  url     = {https://doi.org/10.7759/cureus.79963}
}

@article{US_Preventive,
    author = {US Preventive Services Task Force},
    title = {Screening for Hypertension in Adults: US Preventive Services Task Force Reaffirmation Recommendation Statement},
    journal = {JAMA},
    volume = {325},
    number = {16},
    pages = {1650-1656},
    year = {2021},
    month = {04},
    issn = {0098-7484},
    doi = {10.1001/jama.2021.4987},
    url = {https://doi.org/10.1001/jama.2021.4987},
    eprint = {https://jamanetwork.com/journals/jama/articlepdf/2779190/jama_krist_2021_us_210009_1618595321.57023.pdf},
}

@article{VanDerMee2024LabResultsFormats,
  author  = {van der Mee, F. A. M. and Schaper, F. and Jansen, J. and Bons, J. A. P. and Meex, S. J. R. and Cals, J. W. L.},
  title   = {Enhancing Patient Understanding of Laboratory Test Results: Systematic Review of Presentation Formats and Their Impact on Perception, Decision, Action, and Memory},
  journal = {Journal of Medical Internet Research},
  year    = {2024},
  volume  = {26},
  pages   = {e53993},
  doi     = {10.2196/53993},
  url     = {https://doi.org/10.2196/53993}
}

@article{ESR2023StructuredReportingUpdate,
  author  = {{European Society of Radiology (ESR)}},
  title   = {ESR paper on structured reporting in radiology-update 2023},
  journal = {Insights into Imaging},
  year    = {2023},
  volume  = {14},
  number  = {1},
  pages   = {199},
  doi     = {10.1186/s13244-023-01560-0},
  url     = {https://doi.org/10.1186/s13244-023-01560-0}
}

@article{Petrovskaya2023PortalTestResultsScopingReview,
  author  = {Petrovskaya, O. and Karpman, A. and Schilling, J. and Singh, S. and Wegren, L. and Caine, V. and Kusi-Appiah, E. and Geen, W.},
  title   = {Patient and Health Care Provider Perspectives on Patient Access to Test Results via Web Portals: Scoping Review},
  journal = {Journal of Medical Internet Research},
  year    = {2023},
  volume  = {25},
  pages   = {e43765},
  doi     = {10.2196/43765},
  url     = {https://doi.org/10.2196/43765}
}

@article{Timbrell2024ReferenceIntervalLimitations,
  author  = {Timbrell, N. E.},
  title   = {The Role and Limitations of the Reference Interval Within Clinical Chemistry and Its Reliability for Disease Detection},
  journal = {British Journal of Biomedical Science},
  year    = {2024},
  volume  = {81},
  pages   = {12339},
  doi     = {10.3389/bjbs.2024.12339},
  url     = {https://doi.org/10.3389/bjbs.2024.12339}
}

@inproceedings{yang-etal-2023-data,
    title = "Data Augmentation for Radiology Report Simplification",
    author = "Yang, Ziyu  and
      Cherian, Santhosh  and
      Vucetic, Slobodan",
    editor = "Vlachos, Andreas  and
      Augenstein, Isabelle",
    booktitle = "Findings of the Association for Computational Linguistics: EACL 2023",
    month = may,
    year = "2023",
    address = "Dubrovnik, Croatia",
    publisher = "Association for Computational Linguistics",
    url = "https://aclanthology.org/2023.findings-eacl.144/",
    doi = "10.18653/v1/2023.findings-eacl.144",
    pages = "1922--1932"
}

@article{mimiccxr,
  title={MIMIC-CXR, a de-identified publicly available database of chest radiographs with free-text reports},
  author={Johnson, Alistair EW and Pollard, Tom J and Berkowitz, Seth J and Greenbaum, Nathaniel R and Lungren, Matthew P and Deng, Chih-ying and Mark, Roger G and Horng, Steven},
  journal={Scientific data},
  volume={6},
  number={1},
  pages={317},
  year={2019},
  publisher={Nature Publishing Group UK London}
}

@article{KADHIM2026100895,
title = {Application of generative artificial intelligence to utilize unstructured clinical data for acceleration of inflammatory bowel disease research},
journal = {Med},
volume = {7},
number = {1},
pages = {100895},
year = {2026},
issn = {2666-6340},
doi = {https://doi.org/10.1016/j.medj.2025.100895},
url = {https://www.sciencedirect.com/science/article/pii/S2666634025003228},
author = {Alex Z. Kadhim and Zachary Green and Iman Nazari and Jonathan Baker and Michael George and Ashley Heinson and Bhumita Vadgama and Matt Stammers and Christopher M. Kipps and R. Mark Beattie and James J. Ashton and Sarah Ennis}
}

@Article{bioengineering12111194,
AUTHOR = {Gomez-Cabello, Cesar Abraham and Prabha, Srinivasagam and Haider, Syed Ali and Genovese, Ariana and Collaco, Bernardo G. and Wood, Nadia G. and Bagaria, Sanjay and Forte, Antonio Jorge},
TITLE = {Comparative Evaluation of Advanced Chunking for Retrieval-Augmented Generation in Large Language Models for Clinical Decision Support},
JOURNAL = {Bioengineering},
VOLUME = {12},
YEAR = {2025},
NUMBER = {11},
ARTICLE-NUMBER = {1194},
URL = {https://www.mdpi.com/2306-5354/12/11/1194},
PubMedID = {41301150},
ISSN = {2306-5354},
DOI = {10.3390/bioengineering12111194}
}

@misc{medgemma,
      title={MedGemma Technical Report}, 
      author={Andrew Sellergren and Sahar Kazemzadeh and Tiam Jaroensri and Atilla Kiraly and Madeleine Traverse and Timo Kohlberger and Shawn Xu and Fayaz Jamil and Cían Hughes and Charles Lau and Justin Chen and Fereshteh Mahvar and Liron Yatziv and Tiffany Chen and Bram Sterling and Stefanie Anna Baby and Susanna Maria Baby and Jeremy Lai and Samuel Schmidgall and Lu Yang and Kejia Chen and Per Bjornsson and Shashir Reddy and Ryan Brush and Kenneth Philbrick and Mercy Asiedu and Ines Mezerreg and Howard Hu and Howard Yang and Richa Tiwari and Sunny Jansen and Preeti Singh and Yun Liu and Shekoofeh Azizi and Aishwarya Kamath and Johan Ferret and Shreya Pathak and Nino Vieillard and Ramona Merhej and Sarah Perrin and Tatiana Matejovicova and Alexandre Ramé and Morgane Riviere and Louis Rouillard and Thomas Mesnard and Geoffrey Cideron and Jean-bastien Grill and Sabela Ramos and Edouard Yvinec and Michelle Casbon and Elena Buchatskaya and Jean-Baptiste Alayrac and Dmitry Lepikhin and Vlad Feinberg and Sebastian Borgeaud and Alek Andreev and Cassidy Hardin and Robert Dadashi and Léonard Hussenot and Armand Joulin and Olivier Bachem and Yossi Matias and Katherine Chou and Avinatan Hassidim and Kavi Goel and Clement Farabet and Joelle Barral and Tris Warkentin and Jonathon Shlens and David Fleet and Victor Cotruta and Omar Sanseviero and Gus Martins and Phoebe Kirk and Anand Rao and Shravya Shetty and David F. Steiner and Can Kirmizibayrak and Rory Pilgrim and Daniel Golden and Lin Yang},
      year={2026},
      eprint={2507.05201},
      archivePrefix={arXiv},
      primaryClass={cs.AI},
      url={https://arxiv.org/abs/2507.05201}, 
}

@article{Liss2021GeneralHealthChecks,
  author  = {Liss, David T. and Uchida, Toshiko and Wilkes, Cheryl L. and Radakrishnan, Ankitha and Linder, Jeffrey A.},
  title   = {General Health Checks in Adult Primary Care: A Review},
  journal = {JAMA},
  year    = {2021},
  volume  = {325},
  number  = {22},
  pages   = {2294--2306},
  doi     = {10.1001/jama.2021.6524},
  pmid    = {34100866}
}

@article{10.1145/3458754,
author = {Gu, Yu and Tinn, Robert and Cheng, Hao and Lucas, Michael and Usuyama, Naoto and Liu, Xiaodong and Naumann, Tristan and Gao, Jianfeng and Poon, Hoifung},
title = {Domain-Specific Language Model Pretraining for Biomedical Natural Language Processing},
year = {2021},
issue_date = {January 2022},
publisher = {Association for Computing Machinery},
address = {New York, NY, USA},
volume = {3},
number = {1},
url = {https://doi.org/10.1145/3458754},
doi = {10.1145/3458754},
journal = {ACM Trans. Comput. Healthcare},
month = oct,
articleno = {2},
numpages = {23}
}

@techreport{zipf2013nhanes,
  title       = {National Health and Nutrition Examination Survey:
                 Plan and Operations, 1999--2010},
  author      = {Zipf, George and Chiappa, Michele and Porter, Kathryn S.
                 and Ostchega, Yechiam and Lewis, Brenda G. and
                 Dostal, Jennifer},
  institution = {National Center for Health Statistics},
  series      = {Vital and Health Statistics},
  type        = {Series 1},
  number      = {56},
  address     = {Hyattsville, MD},
  year        = {2013},
  url         = {https://www.cdc.gov/nchs/data/series/sr_01/sr01_056.pdf}
}

@article{lin2026clinically,
  title   = {Clinically Grounded Multi-Agent Artificial Intelligence
             for Preventive Health Management},
  author  = {Lin, Hao and Zhang, Yang and Ye, Dongxin and He, Sicheng
             and Du, Zhaowu and Yu, Yang and Yu, Xiao and Ren, Liping
             and Dong, Nanqing and Hu, Fang and Su, Jinsong and
             Zhang, Jie and Tan, Yun and Zhao, Li},
  journal = {Research Square},
  year    = {2026},
  doi     = {10.21203/rs.3.rs-9150301/v1},
  url     = {https://doi.org/10.21203/rs.3.rs-9150301/v1},
  note    = {Preprint}
}

@inproceedings{mullenbach2021clip,
  title     = {{CLIP}: A Dataset for Extracting Action Items for
               Physicians from Hospital Discharge Notes},
  author    = {Mullenbach, James and Pruksachatkun, Yada and Adler, Sean
               and Seale, Jennifer and Swartz, Jordan and McKelvey, Greg
               and Dai, Hui and Yang, Yi and Sontag, David},
  booktitle = {Proceedings of the 59th Annual Meeting of the Association
               for Computational Linguistics and the 11th International
               Joint Conference on Natural Language Processing
               (Volume 1: Long Papers)},
  pages     = {1365--1378},
  publisher = {Association for Computational Linguistics},
  year      = {2021},
  doi       = {10.18653/v1/2021.acl-long.109},
  url       = {https://aclanthology.org/2021.acl-long.109/}
}

@inproceedings{lau2020followup,
  title     = {Extraction and Analysis of Clinically Important
               Follow-up Recommendations in a Large Radiology Dataset},
  author    = {Lau, Wilson and Payne, Thomas H. and Uzuner, Ozlem
               and Yetisgen, Meliha},
  booktitle = {AMIA Joint Summits on Translational Science Proceedings},
  pages     = {335--344},
  year      = {2020},
  pmid      = {32477653},
  url       = {https://pmc.ncbi.nlm.nih.gov/articles/PMC7233090/}
}

@article{packhauser2022reidentification,
  title   = {Deep Learning-Based Patient Re-identification Is Able to
             Exploit the Biometric Nature of Medical Chest X-Ray Data},
  author  = {Packh{\"a}user, Kai and G{\"u}ndel, Sebastian and
             M{\"u}nster, Nicolas and Syben, Christopher and
             Christlein, Vincent and Maier, Andreas},
  journal = {Scientific Reports},
  volume  = {12},
  number  = {1},
  pages   = {14851},
  year    = {2022},
  doi     = {10.1038/s41598-022-19045-3},
  url     = {https://doi.org/10.1038/s41598-022-19045-3}
}

@article{macdonald2024deidentification,
  title   = {A Method for Efficient De-identification of {DICOM}
             Metadata and Burned-in Pixel Text},
  author  = {Macdonald, Jacob A. and Morgan, Katelyn R. and
             Konkel, Brandon and Abdullah, Kulsoom and Martin, Mark
             and Ennis, Cory and Lo, Joseph Y. and Stroo, Marissa
             and Snyder, Denise C. and Bashir, Mustafa R.},
  journal = {Journal of Imaging Informatics in Medicine},
  volume  = {37},
  number  = {5},
  pages   = {1--7},
  year    = {2024},
  doi     = {10.1007/s10278-024-01098-7},
  url     = {https://doi.org/10.1007/s10278-024-01098-7}
}






\end{document}